\pdfoutput=1

\documentclass[11pt]{article}

\usepackage{multirow}

\usepackage[final]{acl}

\usepackage{booktabs, multirow} 
\usepackage{soul}
\usepackage{xcolor,colortbl} 
\usepackage{changepage,threeparttable} 
\usepackage{longtable}
\usepackage{lscape}
\usepackage{geometry}

\usepackage{listings}  

\usepackage{times}
\usepackage{latexsym}

\usepackage{pgf-pie}

\usepackage[T1]{fontenc}

\usepackage[utf8]{inputenc}

\usepackage{microtype}

\usepackage{inconsolata}

\usepackage{graphicx}

\usepackage{amsmath}
\title{Program Synthesis Dialog Agents for Interactive Decision-Making}

\author{
 \textbf{Matthew Toles\textsuperscript{1}},
 \textbf{Nikhil Balwani\textsuperscript{1,2*}},\\
 \textbf{Rattandeep Singh\textsuperscript{1}},\\
 \textbf{Valentina Giulia Sartori Rodriguez\textsuperscript{1,3}},
 \textbf{Zhou Yu\textsuperscript{1}}
\\
 \textsuperscript{1}Columbia University,
 \textsuperscript{2}Amazon,
 \textsuperscript{3}Sciences Po Paris
\\
 \small{
   \textbf{Correspondence:} \href{mailto:mt3639@columbia.edu}{m.toles@columbia.edu}
 }\\
}

\begin{document}
\maketitle
\begin{abstract}
Many real-world eligibility problems, ranging from medical diagnosis to tax planning, can be mapped to decision problems expressed in natural language, wherein a model must make a binary choice based on the features of the user. 
Large-scale domains such as legal codes or frequently updated funding opportunities render human annotation (e.g., web forms or decision trees) impractical, suggesting a need for agents that can automatically assist in decision-making.
Since relevant information is often only known to the user, it is important that these agents can ask the right questions.
To evaluate this task, we propose BeNYfits, a new benchmark for determining user eligibility for multiple overlapping social benefits opportunities through interactive decision-making.
Our experiments show that current language models struggle with frequent hallucinations, with GPT-4o scoring only 35.7 F1 using a ReAct-style chain-of-thought.
We therefore introduce ProADA, a novel approach that uses program synthesis to assist in decision-making by mapping dialog planning to a code generation problem and using gaps in structured data to determine the best next action.
Our agent, ProADA, improves the F1 score to 56.2 while using nearly the same number of dialog turns.
\renewcommand{\thefootnote}{}
\footnote{* This work completed before joining Amazon}
\renewcommand{\thefootnote}{\arabic{footnote}}
 
\end{abstract}

\section{Introduction}

The improved capabilities of large language models have refocused attention away from traditional benchmarks and towards real-world tasks where automated systems could broadly benefit the public, such as improving access to public services. 
Many such opportunities require determining the user's eligibility based on the user's features and the task at hand, formally referred to as decision problems.
In adaptive decision scenarios, where information is revealed iteratively (e.g., medical diagnosis), one also wishes to minimize the number of queries.
Also, critical information may be user-specific and known only to certain people, meaning that it often \textbf{must} be requested through dialog.
Finally, the diversity of problems in the real world places a premium on whether agents can generalize information gathering and logical reasoning to new domains.

User-facing decision problems have traditionally been solved using hard-coded forms (as in the American tax filing software TurboTax\footnote{https://turbotax.intuit.com/}) or dialog trees (as in video games).
However, hard-coded solutions struggle to generalize or extend to web-scale decision problems; opportunities found by web-scraping, crowd-sourcing, or from extremely large corpora such as national tax codes may be challenging to formalize, let alone decide on in real-time.
Methods to approach adaptive decision problems include multi-armed bandits, reinforcement learning, dynamic programming, and decision trees, but adapting these strategies to online natural language tasks is not trivial.
Recently, large language models have improved on a wide range of related tasks.
However, they are known to struggle with reasoning over long contexts and hallucinate unstated information.

\begin{figure}
    \centering
    \includegraphics[width=1.0\linewidth]{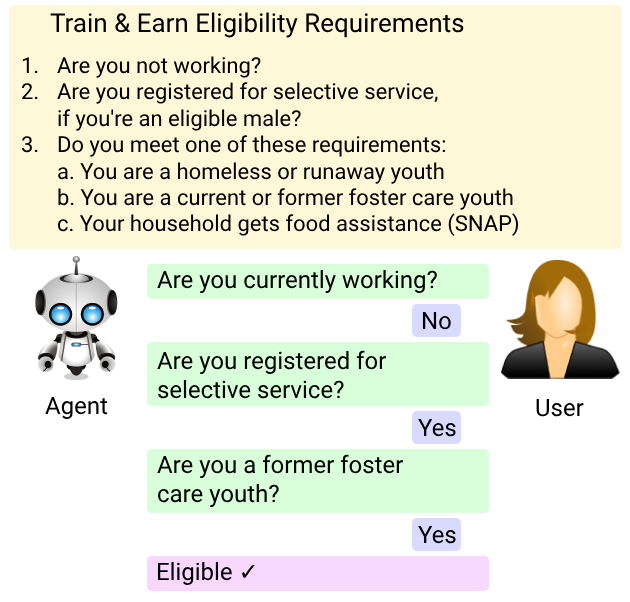}
    \caption{Interactive decision-making dialog loop in BeNYfits. The agent is initialized with opportunity eligibility requirements for the "Train \& Earn" opportunity (simplified). The agent then asks questions to the user until the agent answers \textsc{Yes} to the \textsc{Ready} prompt, at which point it \textsc{Predict}s the user's eligibility. Note that the agent skips requirement 3a because youth cannot register for selective service. Similarly, it skips requirement 3c because it becomes irrelevant if the user is a former foster care youth.}
    \label{fig:dialog-loop}
\end{figure}

To evaluate interactive decision-making, in Section \ref{sec:benyfits} we introduce BeNYfits, a language model agent task for determining user eligibility for real-world public benefits opportunities with overlapping eligibility requirements (Figure \ref{fig:dialog-loop}).
In the single-opportunity scenario, the assistant agent could simply repeat the requirements and ask the user if they qualify.
However, BeNYfits' overlapping requirements present an interesting optimization challenge for dialog planning: how should models ``merge" eligibility requirements to avoid duplicating questions and maximize information gain?
We find that current large language models, including GPT-4o, struggle to perform significantly better than chance at determining user eligibility, suffering from hallucination, poor reasoning under uncertainty, overconfidence, and lost-in-the-middle problems observed in prior work \cite{huang2024survey}.


Given these weaknesses, we introduce a method for an agent that, given a natural language description of the user-facing decision problem, generates a program to request user input conversationally to solve the problem.
Specifically, we construct an agent consisting of a code module, which conducts dialog planning in the form of a Python program, and a dialog module, which asks questions based on the program state.
The agent then uses the dialog module to parse the user's response into structured data.
This approach exploits code generation models' long-range planning and and uncertainty handling to improve task-oriented dialog, as compared to conventional dialog models.
As a key contribution, in Section \ref{sec:proada} we present \underline{Pro}gram Synthesis \underline{A}daptive \underline{D}ecision \underline{A}gent, or ProADA, an agent that, given a natural language policy of a decision problem, generates Python code to structure the decision-making process and request minimal user input to make the correct decision.

\noindent Our main contributions are as follows.\\
\noindent \textbf{1.} A novel agent benchmark for adaptive decision-making in dialog measuring agent accuracy and dialog turn efficiency in helping users determine eligibility for public, real-world opportunities.\\
\noindent \textbf{2.} A general and effective agent for adaptive decision-making in dialog that exploits program synthesis and tool use to plan dialog and adaptively request user information, improving both F1 score and dialog completion speed. 
We will release the model code and maintain a benchmark after publication.

\section{BeNYfits: An Agent Benchmark for Public Benefit Eligibility Decisions} 
\label{sec:benyfits}

The determination of eligibility for many public opportunities, such as tax credits, scholarships, research funding opportunities, business incentives, charities, job listings, and social services, can be reduced to a binary decision problem. 
Since many requirements, such as age and income, overlap between programs, this creates an opportunity for agent assistants to make more efficient and adaptive decisions as compared to traditional methods like static web forms.
At the same time, determinations often require domain-specific knowledge to make accurate determinations, presenting a challenge in natural language understanding.
We present BeNYfits, a benchmark for decision-making on public benefits eligibility.
In BeNYfits, the agent's goal is to help users navigate complex decision-making processes and make a final determination on the user's eligibility based on the dialog in the minimum number of dialog turns.

\subsection{Efficiency, Generalization, and User Experience}

Traditional methods for determining eligibility present several opportunities for improvement by intelligent agent assistants.
For small numbers of opportunities, we might convert natural language requirements into a web form or static chatbot dialog tree serving as an interface for hard-coded checking logic, similar to TurboTax.
However, this approach has several drawbacks.
First, many eligibility requirements are updated without notice, often annually, meaning eligibility performance will degrade over time without ongoing maintenance.
Second, opportunities may be crowdsourced or scraped from the Web dynamically, rendering manual coding impractical in favor of more generalizable, lower-latency solutions, such as language agents.
Furthermore, requirements between similar opportunities frequently overlap, forcing users to answer the same questions repeatedly.
On the other hand, users may waste a lot of time if they discover that they are ineligible very late in the examination process, presenting an opportunity for adaptive decision-making algorithms.
An intelligent agent, however, should adaptively query the user for only the minimum necessary information, saving time, and improving user experience.

In BeNYfits, an agent must interact with a simulated user to determine their household's eligibility for multiple overlapping benefits opportunities based only on the opportunity requirements and conversation with the user.
We define the task as follows. Given a set of opportunities, each with a unique set of eligibility requirements, determine whether the user is eligible for each of them in the minimum number of dialog turns (Figure \ref{fig:dialog-loop}).
We simulate a user by prompting a language model with detailed information about themselves and their household.
Each simulated user is interested in a subset of all opportunities.
Assistant agents possess the natural language eligibility requirements for those opportunities and must determine the simulated user's eligibility by asking a series of questions.
After each dialog turn, the agent determines whether it is ready to make a decision and, if so, outputs a final eligibility prediction for each opportunity.

\subsection{Opportunity Requirements}
\begin{figure}
    \centering
    \includegraphics[width=1.0\linewidth]{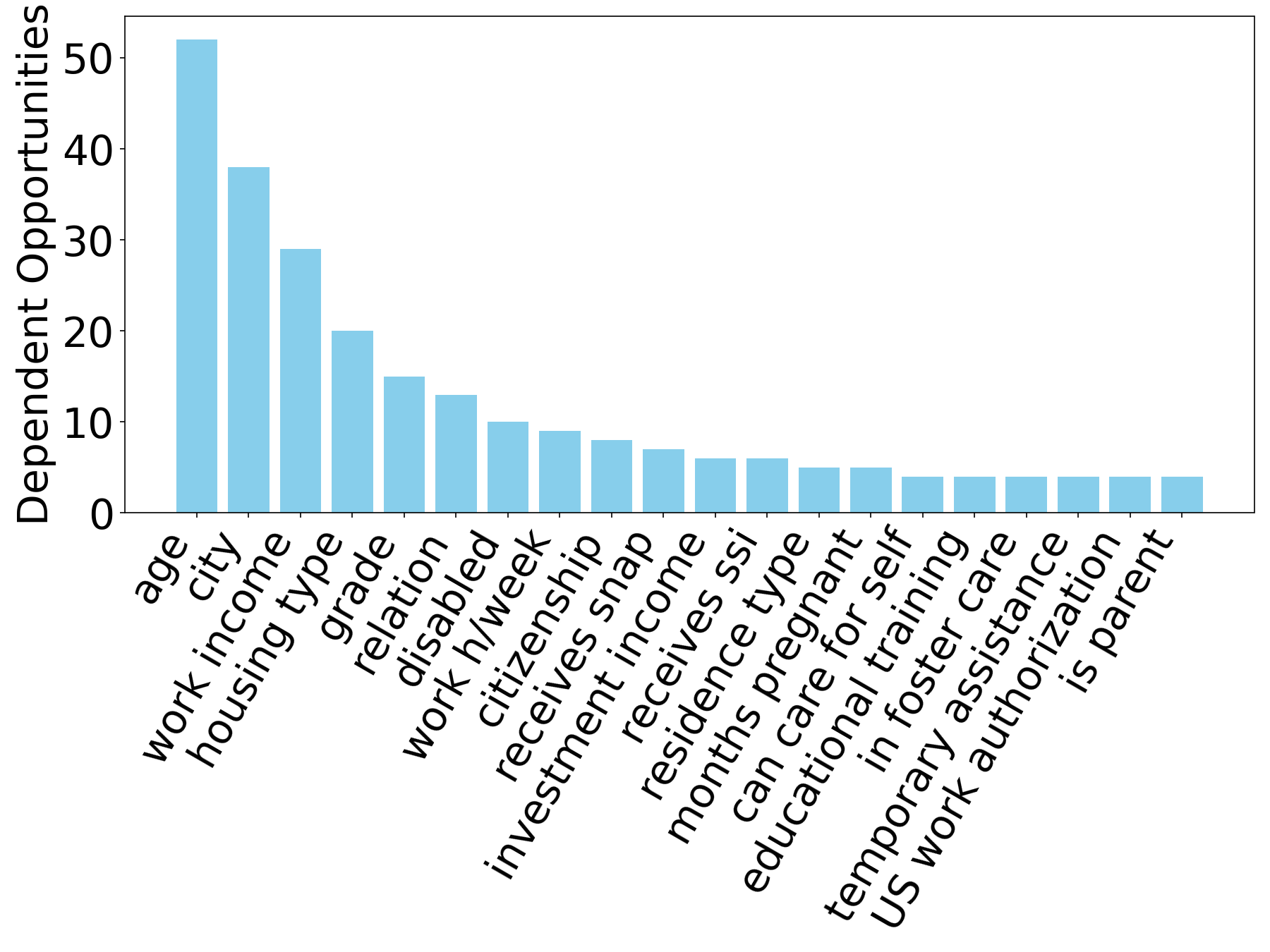}
    \caption{Number of opportunities dependent on each household feature. For example, 53 of 82 programs rely on age to determine eligibility. Top 20 features shown.}
    \label{fig:feature-distribution}
\end{figure}

We source the plain English eligibility requirements for 82 benefits opportunities from NYC Open Data\footnote{https://data.cityofnewyork.us/}. 
We minimally edit requirements to remove ambiguity ("may be eligible"), future expectations ("can commit to"), and dates ("since 2023"). 
Opportunities include tax credits, youth programming, housing, nutrition assistance, healthcare, parental services, and career advancement, among other categories.
Eligibility requirements range from broad (State ID card: all residents age 10+) to extremely specific (Air conditioner subsidy: ten independent requirements).
Opportunities may apply to either the individual or the household as a whole, offering additional logical complexity.
Opportunities depend on 1-18 unique user features each (mean: 4.66 standard deviation: 3.56).
Each user feature appears in 1-52 opportunities (mean: 3.25, standard deviation: 6.66), falling on a long-tailed distribution (Figure \ref{fig:feature-distribution}). 
We define a household as eligible for an opportunity if \textbf{any} of its members are eligible.

\subsection{User Simulation}
For each opportunity, we enumerate relevant user features (age, income, number of dependents, etc.).
We create simulated user households by randomly sampling each feature for each member, with up to 6 members per household.
Features are independently sampled, except when subject to constraints preventing illogical combinations (5-year-old grandparents, adults in foster care, multiple spouses, etc.)
From these structured feature sets, we generate a natural language profile for the household.
We prompt Llama 3.1 70B with the natural language profile to answer questions from the information seeking agent.

\subsection{Eligibility Checker Programs}
To determine the ground truth eligibility of simulated users for specific opportunities, we manually write an eligibility checker Python program for each opportunity based on its plain language requirements.
The eligibility checker takes a simulated user's structured features as input and outputs the user's eligibility for the opportunity.
We take care to avoid Python \textsc{or} and \textsc{and} keywords and other patterns such as list comprehension.
In this way, we ensure that if and only if two households qualify (or fail to qualify) for an opportunity \textbf{for the same reasons}, they cause the eligibility checker to execute the same unique set of lines of code, its trace.

\subsection{Diverse Dataset}

Due to the time and cost of benchmarking new models, we attempt to construct the smallest possible dataset with the most "coverage" of unique traces to qualification or disqualification using input fuzzing.
Each example consists of a simulated user, a subset of opportunities in which they are interested, and the ground truth eligibility for those opportunities.
We first randomly generate 10,000 simulated users, sampling each variable from a distribution chosen to produce a balanced sampling across traces, typically uniformly, except when eligibility is based on thresholds of numeric features.
We then greedily add households to the dataset whose trace through all eligibility checkers execute the most unique lines of code not yet present in the dataset.
Finally, we greedily remove opportunities from simulated users if the user-opportunity trace contributes no unique lines.
We refer to this dataset as the Diverse Dataset, which contains only 56 of the original 10,000 households (305 of 82,000 user-opportunity pairs) but covers the same traces as the full set.
Each household is interested in between 1-10 opportunities, with a mean of 5.4.

\subsection{Representative Dataset}
To model a realistic distribution of potential users, we construct a second Representative Dataset.
Each feature for each user household (e.g., housing type) is independently sampled from distributions derived from 78 different sources.
We use data from New York City when available, but fall back to state- or national-level statistics if necessary.
We assign opportunities to users at random.
The representative contains 25 user households, each interested in 6-19 opportunities, with a mean of 9.8.

\subsection{Dialog Loop}

Agents are provided eligibility requirements and then must determine simulated user eligibility by asking questions, one at a time. 
After each response, the agent is prompted with \textsc{Ready}, where it is asked if it has enough information to determine the eligibility of the user with certainty (Figure \ref{fig:dialog-loop}).
If the agent responds with \textsc{True}, it is prompted to \textsc{Predict} the user's eligibility for each opportunity.
Otherwise, it asks another question.
We limit conversations to average 20 questions per opportunity, to a maximum of 100 questions.

\section{ProADA}
\label{sec:proada}

\begin{figure*}
    \centering
    \includegraphics[width=1.0\linewidth]{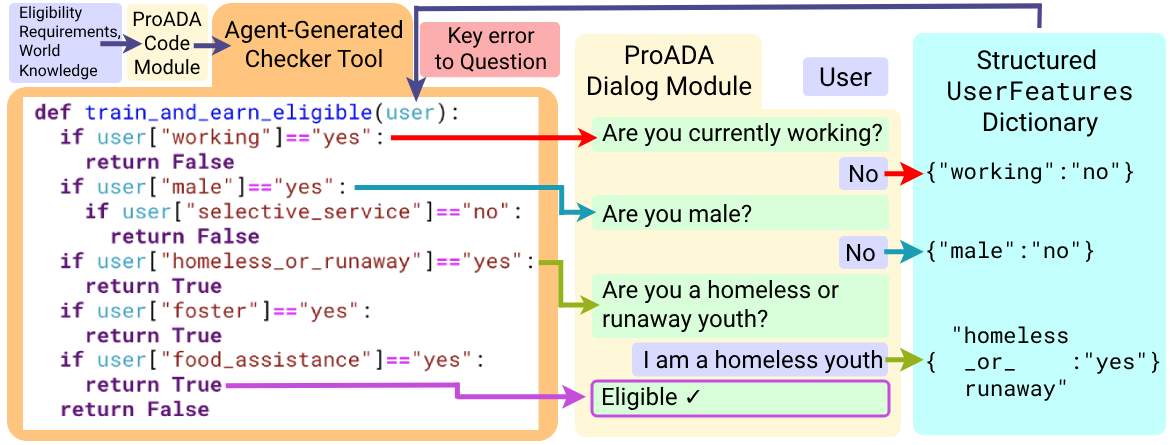}
    \caption{ProADA architecture. ProADA consists of the checker tool created by the code generation module (left) and the dialog module (center). The checker tool is a Python function that determines user eligibility from a structured user representation dictionary (right). The ProADA dialog module acts as an interface between the checker tool and the user. On each dialog turn, the agent runs the checker tool on the user dictionary, which is initially empty. On a key error, the dialog module fills in a single key-value pair by asking a user a question and converting the answer to a value consistent with the checker tool logic. The dialog ends once a value is returned by the checker tool for every opportunity.}
    \label{fig:ProADA}
\end{figure*}

To solve interactive decision problems, we propose a Program Synthesis Adaptive Decision Agent, or ProADA, which uses agent-created Python tools as reasoning aids for adaptive decision problems in dialog.
State-of-the-art code generation models often generate code that involves a dozen variables \cite{wan2024deep}, yet the models suffer from basic reasoning errors and hallucinations when working in natural language.
By offloading dialog planning and memory into static Python code, ProADA achieves the flexibility and usability of natural language while leveraging the long-range planning and reasoning of program synthesis.
ProADA consists of a code generation module and a dialog module.
The code generation module creates one Python \textsc{Decide} tool per opportunity, formalizing the logic of the decision problem and deciding the result.
The dialog module serves as an interface between the user and the \textsc{Decide} tool, asking questions and storing answers in a structured form (Figure \ref{fig:ProADA}).

To best explain ProADA, we instantiate it in the context of our proposed BeNYfits benchmark.
Before starting a dialog, ProADA uses the code generation model to convert the eligibility requirements in natural language into a Python \textsc{Decide} Checker tool used by the agent (Figure \ref{fig:ProADA}).  
\textsc{Decide} is a Python function that takes a \texttt{UserFeatures} dictionary containing known user properties (e.g., \texttt{``homeless\_or\_runaway"}) as input and outputs the household eligibility.
For each key used to access \texttt{UserFeatures}, the code generation model defines a type (\texttt{int}, \texttt{float}, etc.) constraint, or a list of string choices that the feature can take.
At the start of the dialog, the agent runs the \textsc{Decide} tool, passing in an empty \texttt{UserFeatures} dictionary, since it knows nothing about the user yet.
An empty dictionary would normally cause a key error, which we exploit by wrapping \textsc{Decide} in a try/except block.
In an exception, the agent passes the offending key, relevant code, eligibility requirements, and dialog history to the dialog module.
The dialog module constructs a question seeking the necessary information (``Are you a homeless or a runaway youth?") and presents it to the user.
The dialog module then converts the user's response (``I am a homeless youth") into a valid value according to the predetermined constraint using constrained generation, storing the key-value pair (\texttt{``homeless\_or\_runaway": ``yes")} in \texttt{UserFeatures}.
In the case that the user's response cannot be mapped to a valid value, ProADA will invoke the \textsc{Clarity} module, which will spend up to three turns attempting to clarify the user's response (additional details in Appendix \ref{appendix:key-error}).
This module allows ProADA to recover from unhappy dialog paths and incoherent user input.
Finally, the agent repeats running \textsc{Decide} with the updated \texttt{UserFeatures} dictionary until it returns a value.

\section{Experimental Setup}
\label{sec:experiments}

As baselines, we choose Llama 3.1 Instruct 8B and 70B, as well as GPT-4o. 
In direct prompting, we instruct these models to assess readiness and generate questions at each step in the dialog loop, and finally to predict eligibility.
We also assess prompting models to conduct ReAct-style chain-of-thought before each step \cite{yao2023react}.
During \textsc{Readly} and \textsc{Decide}, we use constrained decoding to ensure ProADA and baseline models generate a valid output.
For ProADA, we choose the same models for the dialog module and always use GPT-4o for the code generation module.
We choose Llama 3.1 Instruct 70B to implement our simulated user for all experiments, in order to reduce the hallucinations that we observed in smaller 8B parameter models.
To reduce memory usage, we use 4-bit quantization on all 70B-parameter models.
We report three trials for Llama 8B, Llama 70B ReAct, and ProADA models, and one trial for others due to resource constraints.
To measure human performance on this task, two expert authors performed the role of the agent on 83 user-opportunity pairs from the Diverse Dataset, achieving 85.6\% accuracy. 
Upon review, we find all inaccuracies are due to human error rather than unfaithful simulated user responses, suggesting a performance ceiling near 100\% on this benchmark.

\section{Experimental Results}
\label{sec:results}

\begin{figure*}[h]
    \centering
    \includegraphics[width=1.0\linewidth]{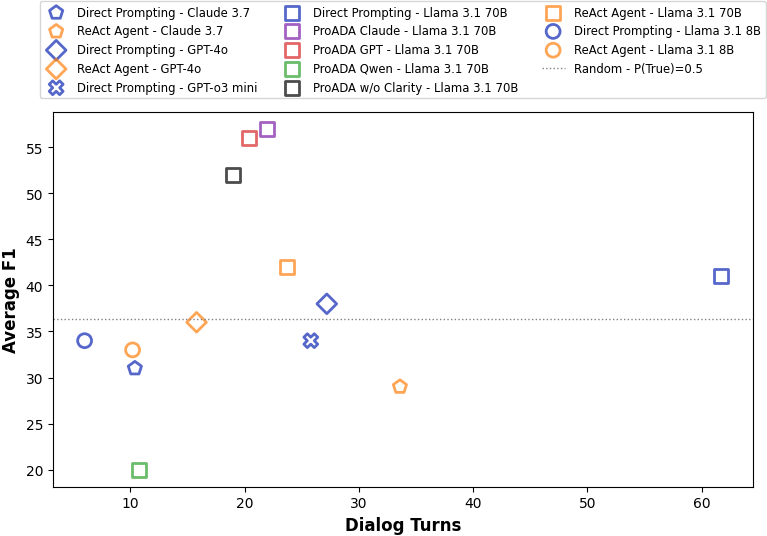}
    \caption{Average of Representative and Diverse dataset F1 vs. dialog turns to completion for ProADA and baseline models. Legend follows the format Strategy (Code Model) - Dialog Model.}
    \label{fig:main}
\end{figure*}

Because our datasets are unbalanced (Diverse: 47.9\% positive, Representative: 15.5\%), we choose micro F1 as our primary accuracy metric. 
Let F1 and $T$ be the average F1 score and the number of turns across both datasets, respectively. 
To reward efficient questioning, we define a turn-weighted F1 score as:

\begin{equation}
    \label{eq:twf1}
    \text{Turn-Weighted F1} = \frac{100 \cdot \text{F1}}{T / 100 + 1}
\end{equation}

We choose this metric because it increases monotonically from 0 to 1 with respect to F1 and decreases monotonically with respect to turns. The inclusion of the $+1$ term prevents small changes in $T$ from drastically impacting turn-weighted F2 when $T$ is small. Holding F1 constant, turn-weighted F1 will range from F1 to $0.5 \cdot \text{F1}$ as $T$ increases. Because $T\leq100$, we normalize it by 100.

We find that our method, ProADA, outperforms all others by a significant margin with a turn-weighted F1 score of 46.7 (56.2 F1, 20.4 turns, Figure \ref{fig:main}).
The turn-weighted F1 score drops to 43.5 without the \textsc{Clarity} module.
However, ProADA still surpasses the next best strategy, Llama 3.1 70B + ReAct (34.1) and GPT-4o + direct prompting (29.9), despite using a dialog model with many times fewer parameters. 
GPT-4o achieves a relatively high average F1 score (40.8), but uses 27.2 turns per dialog, while Llama 3.1 8B appears to terminate prematurely after only 6.0 turns, achieving only 33.6 F1.
Claude 3.7 slightly outperforms GPT-4o when serving as the ProADA code module (47.1 vs 46.7 Turn-Weighted F1), likely due to its internal inference-time scaling.
However, GPT-4o outperforms in the direct prompting and ReAct conditions, leading us to choose it as our primary baseline.
Dialog completion speed varies widely across models and strategies, with Llama 3.1 70B + direct prompting frequently hitting the turn limit without terminating, resulting in an average turn count of 81.7. 
Program synthesis guidance, and to a lesser degree ReAct prompting, appear to moderate the number of turns needed without negatively impacting F1 score.
We see significantly higher accuracy on the Diverse Dataset compared to the Representative Dataset, possibly because its examples contain roughly half as many opportunities per user.

\subsection{User Study}

We conducted a user survey n=58 interactions.
Each user selected 5 opportunities in which they are personally interested. Half of participants were assigned each to use either ProADA or GPT-4o + ReAct. 
Users then rated the accuracy and subjective usability on a 5-point likert scale, in accordance with the NASA Task Load Index \cite{hart2006nasa}.
We find that ProADA achieves better usability scores in metrics as well as significantly higher accuracy and program discovery with real users (Table \ref{tab:user}).

\begin{table*}[!htp]\centering
\small
\begin{tabular}{lrrrrrrrrrr}\toprule
\textbf{} &\multicolumn{2}{c}{\textbf{Model}} &\multicolumn{2}{c}{\textbf{Diverse}} &\multicolumn{2}{c}{\textbf{Representative}} &\multicolumn{2}{c}{\textbf{Average}} &\textbf{Weighted} \\\cmidrule{2-10}
\textbf{Strategy} &\textbf{Dialog} &\textbf{Code} &\textbf{F1 ↑} &\textbf{Turns ↓} &\textbf{F1 ↑} &\textbf{Turns ↓} &\textbf{F1 ↑} &\textbf{Turns ↓} &\textbf{F1 ↑} \\\midrule
Direct Prompting &Llama 3.1 8B &GPT-4o &44.6 &5.5 &22.5 &6.4 &33.6 &6.0 &31.7 \\
&Llama 3.1 70B &GPT-4o &54.3 &41.6 &27.2 &81.7 &40.8 &61.7 &25.2 \\
&GPT-4o &GPT-4o &31.2 &16.8 &45.0 &37.6 &38.1 &27.2 &29.9 \\
&GPT-o3 mini &GPT-4o &52.9 &20.3 &15.6 &31.2 &34.2 &25.8 &27.2 \\
&Claude 3.7 &GPT-4o &43.7 &11.9 &17.9 &8.9 &30.8 &10.4 &27.9 \\
ReAct Agent &Llama 3.1 8B &GPT-4o &40.4 &9.5 &26.1 &10.9 &33.2 &10.2 &30.2 \\
&Llama 3.1 70B &GPT-4o &58.3 &9.5 &26.1 &37.9 &42.2 &23.7 &34.1 \\
&GPT-4o &GPT-4o &50.4 &18.4 &20.9 &13.3 &35.7 &15.8 &30.8 \\
&Claude 3.7 &GPT-4o &41.3 &38.7 &16.4 &28.4 &28.8 &33.6 &21.6 \\
Random &P(True=0.5) &- &23.6 &0 &48.9 &0 &36.3 &0 &36.3 \\
ProADA (ours) &Llama 3.1 70B &GPT-4o &\textbf{58.7} &18.8 &53.7 &22.1 &56.2 &20.4 &46.7 \\
&Llama 3.1 70B &Claude 3.7 &57.0 &21.5 &\textbf{58.0} &22.4 &\textbf{57.5} &22.0 &\textbf{47.1} \\
&Llama 3.1 70B &Qwen 32B &21.9 &10.1 &17.9 &11.5 &19.9 &10.8 &17.9 \\
ProADA w/o Clarity &Llama 3.1 70B &GPT-4o &54.0 &18.2 &49.7 &19.8 &51.8 &19.0 &43.5 \\
\bottomrule
\end{tabular}
\caption{F1 score and dialog turns to completion for ProADA and baseline models. Average is averaged across Diverse and Representative Datasets. Weighted F1 score uses equation \eqref{eq:twf1}.}\label{tab:results}
\end{table*}

\section{Failure Analysis}
\label{sec:failure}

We observe multiple distinct types of errors that contribute to poor reasoning and inefficient dialog. 
Program synthesis-guided dialog reduces errors overall, but introduces unique failure modes associated with code generation.
However, several failure modes persist across all strategies, indicating core weaknesses in foundational model reasoning ability.

\noindent \textbf{Suggestibility:} Models suffer from hallucination prompted by implications in eligibility requirements.
For example, when prompted with a child care program, models ask for the child's age without checking whether the household contains any children to begin with.

\noindent \textbf{Domain knowledge \& edge cases:} Models fail to account for edge cases, such as 17-year-olds with work income or adult dependents.

\subsection{Baseline Behavior}
Although it is difficult to confidently attribute final predictions to specific mistakes in black-box models during question generation, we observe several flawed reasoning patterns when using direct and ReAct prompting:

\noindent \textbf{Hallucination:} Baseline models frequently return \textsc{True} in \textsc{Ready} before collecting all relevant information, implying either a logical reasoning failure or an internal hallucination of relevant facts. 

\noindent \textbf{Hyperspecificity:} Models ask needlessly specific questions ("Is your total annual income below \$69,900?") when a more general question ("What is your total annual investment income?") would produce information useful elsewhere, resulting in superfluous dialog turns.

\noindent \textbf{Repetition:} Baseline models get stuck in loops, asking slight variations of the same question. 

\noindent \textbf{Multi-Member Households:} Baseline models often inquire only about the user, rather than all members of the family, despite being specifically prompted to do so.
They rarely ask for the family size or composition when eligibility is determined at the individual level, substantially reducing recall.

\noindent \textbf{Conflating Users:} Baseline models often conflate household members or fail to specify which member they are asking about. 

\subsection{ProADA Behavior}
Program synthesis-guided dialog introduces several distinct new failure modes:

\noindent \textbf{Code Generation:} Logical or domain-specific reasoning errors can create flawed code that propagates errors through subsequent conversations. 

\noindent \textbf{Code to Question:} Although the code generated for the \textsc{Decide} tool usually represents multiple family members correctly as a list, the dialog module struggles to track and specify which member is being discussed at any time. 
Interestingly, we observe improved performance when users provide the names of their family members.

To investigate bias arising from the mapping of user responses to structured data, we conduct the following study: We analyze 17 questions generated by ProADA and provide 6 perturbed versions of the same answer. 
Across all 102 trials, we find that overall, user responses are rarely mapped incorrectly, in only 2\% (n=2) of cases.

\begin{adjustwidth}{-2.5 cm}{-2.5 cm}\centering
\begin{threeparttable}[!htb]\centering
\small
\begin{tabular}{lrrr}\toprule
\textbf{} &\textbf{ProADA} &\textbf{GPT-4o + React} \\\midrule
\textbf{Accuracy ↑} &\textbf{81.3\%\textsuperscript{†}} &42.1\% \\
\textbf{Discovered Opportunity ↑} &\textbf{80.0\%\textsuperscript{†}} &50.0\% \\
\midrule
\textbf{Clarity ↑} &\textbf{4.80\textsuperscript{†}} &4.14 \\
\textbf{Ease of Use ↑} &\textbf{4.53\textsuperscript{*}} &3.96 \\
\textbf{Could Save Time ↑} &\textbf{4.47\textsuperscript{†}} &3.39 \\
\textbf{Question Appropriateness ↑} &\textbf{4.43\textsuperscript{†}} &3.54 \\
\textbf{Task Complexity ↓} &\textbf{1.73\textsuperscript{†}} &1.89 \\
\textbf{Workload ↓} &\textbf{1.33\textsuperscript{†}} &2.04 \\
\textbf{Irritation ↓} &\textbf{1.50\textsuperscript{†}} &2.71 \\
\bottomrule
\end{tabular}
\caption{User-reported accuracy, usability, and frequency of discovery of at least one opportunity for which the user is eligible. * p<0.05, † p<0.01}
\label{tab:user}
\end{threeparttable}
\end{adjustwidth}


\begin{figure}
    \centering
    \includegraphics[width=0.9\linewidth]{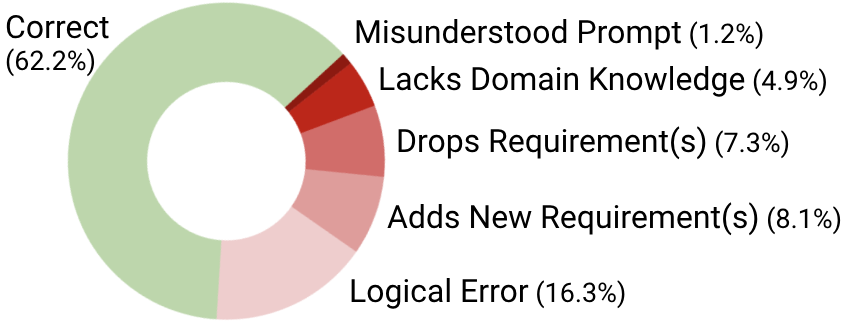}
    \caption{ProADA program synthesis errors}
    \label{fig:pie}
\end{figure}

\subsection{Errors by Simulated Users}
Authors annotated 61 simulated user responses for faithfulness to the user profile, finding 60 (98.4\%) of questions are answered faithfully (Figure \ref{fig:pie}).
The simulated user tends towards verbosity, providing additional unrequested information in 5 cases (8.2\%).
We find unnatural but faithful responses in 2 cases (3. 3\%), indicating that the frequency of errors due to simulated user misbehavior is low.
In qualitative probing, we find that the simulated user can respond accurately to diverse questions up to two hops (e.g., "How many children do you have under the age of 5?").
Sufficiently complex queries or those with more than two hops tend to cause the simulated user to respond that it cannot answer the question, but we rarely observe models generating such questions in our experiments.

\section{Discussion}
\label{sec:discussion}

Program-synthesis-guided dialog improves accuracy in adaptive decision problems while reducing the number of dialog turns needed.
This provides multiple benefits by exposing the agent's reasoning process in a human-readable format.
Agent decisions become more transparent and consistent, improving interpretability, and enabling several avenues for further improvements.

Since the Python tool only needs to be created once, we can use a stronger model for program synthesis without incurring significantly increased inference costs or latency.
Then, by replacing the \textsc{Ready} and \textsc{Predict} language model calls in the dialog loop with simple Python functions, we reduce the number of language model calls by over 50\%.
Unlike in black-box models where we observe disparate behavior based on surface form variation, especially in out-of-distribution contexts, our technique forces the agent to behave consistently across users.
As a form of prompt transformation, this may also reduce the susceptibility of public-facing agents to jailbreak \cite{peng2024jailbreaking}.

This white-box reasoning approach enables transparency into model failure modes and clear opportunities for future work.
On the dialog side, we observe that LLMs often get stuck in dialog loops after making false assumptions about the user.
In code generation, we observe a significant portion of errors are due to inaccurate code generation (Figure \ref{fig:pie}) related to domain expertise.
Specifically, coding agents could search for domain knowledge (e.g., the tax-purposes definition of a dependent in the state of New York) before conducting program synthesis. 
Strategies in self-correction and theory of mind may be applicable here.
Although we generate code automatically in this work, the code may be checked manually or with software tools to ensure correctness before deployment.
Unlike black-box models, program synthesis-guided models like ProADA may also be subject to unit tests to ensure code quality.

AI faces increasing regulation, especially in public services or where systemic bias may disenfranchise certain groups, such as credit offerings.
In certain scenarios, providers are required to prove that their models are unbiased or to provide a human-readable basis for any given AI decision.
Although questions are generated neurally, eligibility decisions are made with static code that can be automatically traced to produce a rationale.
However, we note that the parsing of user utterances into structured data may still introduce bias.

Many opportunities in BeNYfits and other public opportunities are contingent on sensitive personal information, such as income, substance abuse, domestic violence, and being HIV positive.
By limiting closed-source model use only to program synthesis, solutions like ProADA avoid leaking user data to commercial entities while harnessing their models' advanced reasoning.


ProADA represents a reverse of the traditional tool-use paradigm in which language models call tools by generating special tokens.
Instead, our agent creates a tool which in turn calls the language model.
Future work may explore more sophisticated agent-tool relationships.

\section{Related Work}
\label{sec:related_work}

Many dialog agent tasks have been proposed, including offline task-oriented dialog \cite{andreas2020task} \cite{budzianowski2018multiwoz} and online user simulations using real humans or LM agents as responders \cite{gur2018user} \cite{he2018decoupling}. 
Question generation is a related task where agents seek information relevant to a downstream task, such as user intent \cite{min2020ambigqa}, relevant facts \cite{toles2023good}, or user preferences \cite{li2023eliciting}.
Some task-oriented dialog datasets focus on clarification and information seeking, such as \citet{zhang2023groundialog}.
However, datasets such as ShARC \cite{saeidi2018interpretation} and ClariT \cite{feng2023towards} only require "yes" or "no" questions.
BeNYfits expands on these works by adding a highly realistic, multi-turn dialog agent task requiring logical reasoning and domain-specific knowledge.
Similar tasks include MediQ \cite{li2024mediq}, which benchmarks medical diagnosis through dialog, and ClarQ-LLM \cite{gan2024clarq}, which focuses on discovering hidden information while playing an adventurer.
In comparison, BeNYfits focuses on logically reasoning legalistic tasks to reach a binary prediction.

Many works on tool-use have equipped language models with a code interpreter \cite{gupta2023visual} \cite{shen2024hugginggpt}, though fewer have specifically studied tool creation, e.g., \citet{qian2023creator}.
Several prior works have established the efficacy of code generation in dialog systems. 
\citet{chiu2023symbolic} propose grounding in code generated based on partner utterances and using symbolic planning to reason over the code. 
\citet{suris2023vipergpt} find code translations an effective intermediate representation for natural language questions.
\citet{nguyen2024dynasaur} create an LLM agent framework for dynamically creating and composing subtask actions based on code.
To the best of our knowledge, no other code generation-based approaches have been proposed for question generation in dialog. 

\section{Conclusion}
\label{sec:conclusion}


We present a strong tool-augmented method to solve interactive decision-making in dialogs and a novel and realistic benchmark for measuring decision-problem accuracy and dialog efficiency.
Our method ameliorates memory and planning issues by converting key information in user utterances into structured key-value pairs to improve reasoning, latency, and cost by offloading computations onto an agent-created Python tool.
Such structured coding support overcomes many problems of pure LLM baselines such as hallucination of missing information, lack of object tracking, being over-confident, etc. 
Ultimately, our proposed method achieved an F1 score of 56.2 (compared to at most 42.2 for the baselines) while reducing the dialog turns needed by 13.9\% compared to the next best agent, raising hopes for reducing user burden and increasing access to public opportunities using language models.

\section{Limitations}

The eligibility requirements for this benchmark were derived from plain English summaries rather than official documents.
Requirements for some opportunities omit details present in more complete sources.

Although our dataset includes numerous state and federal level opportunities, suggesting broader applicability to many US-oriented applications, we agree that NYC Open Data may not generalize to other regions and contexts. We will add the following to the limitations section:

We note that, due to its size and budget, New York City and State provide a wider range of services than most other municipalities around the world. For example, New York State’s right to shelter drives many of the 18 housing-related opportunities in the dataset. Other domains, especially those requiring subjective evaluation (e.g., merit-based opportunities) present additional challenges, as eligibility cannot necessarily be determined consistently based on self-reported user information.

The population data used to construct the Representative Dataset were collected from numerous independent sources.
Some features were not available, such as the percentage of people currently struggling to pay their electricity bill.
In such cases, we make estimates based on the most similar available data.
At the same time, features are each collected from disparate sources, rather than from a single census, so our dataset is unable to express accurate correlations between related features.
Users of our dataset should be aware of these limitations.

Because our evaluation method weights the F1 score against dialog turns, complex, multi-hop queries are weighted the same as simple yes or no questions.
However, in practice, we rarely observe complex queries.
The trade-offs of question complexity, length, and user burden may be addressed in future work.

\section{Ethical Considerations}

Empirically, we observe that model-generated code in this study does not contain harmful side effects.
However, it is always safer to run untrusted code in a sandboxed environment like Docker.

Introducing AI models into the social benefits system poses risks of false determinations and inequitable user experiences.
We encourage stakeholders to use AI to increase accessibility to public opportunities, but to avoid using them as the final determiner in any step due to the harm caused by errors.
Similarly, user-facing deployments should consider the relative harm of false acceptances versus false refusals and calibrate their models accordingly.

\subsubsection*{Acknowledgments}
This material is based upon work supported by the National Science Foundation Graduate Research Fellowship under Grant No. (DGE-2036197). 
We thank Nick Deas and Zachary Horvitz for their helpful feedback on early drafts. 

\bibliography{custom}

\appendix

\section{Structured Data Mapping}

To investigate bias arising from the mapping of user responses to structured data, we conduct the following study: 
We analyze 17 questions generated by ProADA and provide 6 perturbed versions of the same answer. 
For example, to the question “What is the number of people living in your household?”, we provide the following perturbations: 

\noindent\textbf{Numeric}: 1

\noindent\textbf{Text}: one

\noindent\textbf{Verbose}: There is only one person in my household

\noindent\textbf{Multi-hop}: I don’t live with anyone else

\noindent\textbf{Misspelled}: onee

\noindent\textbf{Extraneous Info}: One but I have a dog

Across all 102 trials, we find that overall, user responses are mapped incorrectly only 2\% (n=2) of the time. Inaccurate mapping occurred only once each in multi-hop and extraneous info perturbations. Otherwise, we see ProADA attempt to clarify user responses in 6\% (n=1) of verbose perturbations and 23\% (n=4) misspelling perturbations, which we consider desired behavior. We believe two factors contribute to the low level of bias observed: Firstly, eligibility requirements in BeNYfits are objective, leaving little room for biased interpretation. Secondly, ProADA generates primarily yes/no (65\%, n=11) and numeric questions (35\%, n=6), which are generally straightforward to parse.

\newpage 
\onecolumn
\begin{landscape}
\section{Detailed Results}
\begin{table}[!htp]\centering
\small
\begin{tabular}{lrrrrrrrrrrrrrrr}\toprule
\textbf{} &\textbf{} &\multicolumn{5}{c}{\textbf{Diversity}} &\multicolumn{5}{c}{\textbf{Representative}} &\multicolumn{3}{c}{\textbf{Average}} \\\cmidrule{3-15}
\textbf{Strategy} &\textbf{Model} &\textbf{Accuracy} &\textbf{Recall} &\textbf{Precision} &\textbf{F1} &\textbf{Turns} &\textbf{Accuracy} &\textbf{Recall} &\textbf{Precision} &\textbf{F1} &\textbf{Turns} &\textbf{F1} &\textbf{Turns} &\textbf{TW F1} \\\midrule
Direct Prompting &Llama 3.1 8B &51.5 &37.7 &49.1 &42.6 &8.92 &63.8 &41.7 &19.0 &26.1 &5.01 &34.4 &7.0 &32.1 \\
& &51.5 &37.7 &49.1 &42.6 &3.85 &65.1 &22.2 &12.9 &16.3 &10.00 &29.5 &6.9 &27.6 \\
& &55.7 &43.8 &54.7 &48.7 &3.69 &66.8 &36.1 &19.1 &25.0 &4.32 &36.8 &4.0 &35.4 \\
& &52.9 &39.7 &51.0 &44.6 &5.49 &65.2 &33.3 &17.0 &22.5 &6.44 &33.6 &6.0 &31.7 \\
&Llama 3.1 70B &52.1 &53.4 &50.0 &51.7 &46.75 &66.8 &52.8 &23.8 &32.8 &76.08 &42.2 &61.4 &26.1 \\
& &53.8 &63.0 &51.4 &56.6 &37.05 &66.0 &41.7 &20.3 &27.3 &84.08 &41.9 &60.6 &26.1 \\
& &53.4 &58.9 &51.2 &54.8 &41.07 &62.6 &33.3 &15.8 &21.4 &85.08 &38.1 &63.1 &23.4 \\
& &53.1 &58.4 &50.9 &54.3 &41.62 &65.1 &42.6 &19.9 &27.2 &81.75 &40.8 &61.7 &25.2 \\
&GPT-4o &77.4 &33.3 &29.3 &31.2 &16.75 &55.1 &38.4 &54.4 &45.0 &37.64 &38.1 &27.2 &29.9 \\
&GPT-o3 mini &82.6 &63.9 &45.1 &52.9 &20.32 &77.0 &13.9 &17.9 &15.6 &31.20 &34.2 &25.8 &27.2 \\
&Claude 3.7 &54.4 &37.0 &53.5 &43.7 &11.91 &76.6 &16.7 &19.4 &17.9 &8.88 &30.8 &10.4 &27.9 \\
ReAct Agent &Llama 3.1 8B &56.2 &41.7 &15.5 &22.6 &13.96 &56.2 &41.7 &15.5 &22.6 &13.96 &22.6 &14.0 &19.8 \\
& &50.8 &49.3 &48.6 &49.0 &6.07 &56.6 &47.2 &17.0 &25.0 &8.52 &37.0 &7.3 &34.5 \\
& &52.1 &49.3 &50.0 &49.7 &8.38 &63.4 &52.8 &21.6 &30.6 &10.36 &40.2 &9.4 &36.7 \\
& &53.0 &46.8 &38.0 &40.4 &9.47 &58.7 &47.2 &18.0 &26.1 &10.95 &33.2 &10.2 &30.2 \\
&Llama 3.1 70B &57.4 &62.3 &54.8 &58.3 &9.53 &63.8 &41.7 &19.0 &26.1 &37.92 &42.2 &23.7 &34.1 \\
&GPT-4o &56.7 &45.9 &55.8 &50.4 &18.41 &71.1 &25.0 &18.0 &20.9 &13.28 &35.7 &15.8 &30.8 \\
&Claude 3.7 &57.6 &31.0 &61.6 &41.3 &38.66 &77.9 &13.9 &20.0 &16.4 &28.44 &28.8 &33.6 &21.6 \\
ProADA w/o Clarity &Llama 3.1 8B &62.6 &41.8 &67.8 &51.7 &16.58 &75.7 &58.3 &33.3 &42.4 &22.96 &47.1 &19.8 &39.3 \\
& &61.6 &43.8 &64.6 &52.2 &18.14 &78.7 &61.1 &37.9 &46.8 &27.96 &49.5 &23.1 &40.2 \\
& &59.2 &39.0 &62.0 &47.9 &15.58 &76.2 &44.4 &30.8 &36.4 &19.24 &42.1 &17.4 &35.9 \\
& &61.2 &41.6 &64.8 &50.6 &16.77 &76.9 &54.6 &34.0 &41.9 &23.39 &46.2 &20.1 &38.5 \\
&Llama 3.1 70B &64.9 &49.3 &68.6 &57.4 &18.83 &86.0 &72.2 &53.1 &61.2 &19.12 &59.3 &19.0 &49.8 \\
& &63.0 &43.8 &67.4 &53.1 &18.55 &83.0 &61.1 &45.8 &52.4 &20.20 &52.7 &19.4 &44.2 \\
& &61.5 &42.5 &65.3 &51.5 &17.28 &78.3 &38.9 &32.6 &35.4 &20.16 &43.4 &18.7 &36.6 \\
& &63.1 &45.2 &67.1 &54.0 &18.22 &82.4 &57.4 &43.8 &49.7 &19.83 &51.8 &19.0 &43.5 \\
&GPT-4o &63.3 &39.0 &71.3 &50.4 &15.73 &86.8 &66.7 &55.8 &60.8 &17.20 &55.6 &16.5 &47.7 \\
ProADA &Llama 3.1 70B &70.2 &53.4 &77.2 &63.2 &20.32 &82.6 &63.9 &45.1 &52.9 &24.20 &58.0 &22.3 &47.5 \\
& &66.6 &54.8 &69.0 &61.1 &20.38 &84.7 &55.6 &50.0 &52.6 &24.80 &56.9 &22.6 &46.4 \\
& &67.2 &52.1 &71.7 &60.3 &18.75 &80.0 &61.1 &40.0 &48.4 &22.04 &54.3 &20.4 &45.1 \\
& &66.8 &49.8 &72.3 &58.7 &18.79 &83.5 &61.8 &47.7 &53.7 &22.06 &56.2 &20.4 &46.7 \\
&Claude 3.7 (code) &67.9 &44.5 &79.3 &57.0 &21.5 &87.7 &55.6 &60.6 &58.0 &22.44 &57.5 &22.0 &47.1 \\
&Qwen 32B (code) &53.1 &13.7 &54.1 &21.9 &10.1 &76.6 &16.7 &19.4 &17.9 &11.52 &19.9 &10.8 &17.9 \\
\bottomrule
\end{tabular}
\caption{Complete BeNYfits results}\label{tab: }
\end{table}
\end{landscape}

\newpage

\twocolumn

\section{List of Prompts}

\subsection{``Are Benefits Ready?'' Prompt}

Eligibility requirements: \{eligibility\_requirements\}. 

Is the information sufficient to determine whether any member of the user's household is eligible for all programs? Answer only in one word True or False.

\subsection{``Predict Benefits Eligibility'' Prompt}

Eligibility: \{eligibility\_requirements\}. 

Predict the programs for which any member of the user's household is eligible. Return only a boolean array of length \{num\_programs\}, e.g. \{example\_array\}, where the value at index `i` is true iff the user is eligible for program `i`. Only return the array. Do not return anything else in the response. If a user's eligibility is unclear, make your best guess.

\subsection{``Ask a Clarifying Question'' Prompt}

Eligibility: \{eligibility\_requirements\}. 

Ask a clarifying question that will help you determine if any member of the user's household is eligible for benefits as efficiently as possible. Only ask about one fact at a time.

\subsection{``Predict Benefits Eligibility'' for CoT Prompt}

Eligibility requirements: \{eligibility\_requirements\}. 

Is the information sufficient to determine whether any member of the user's household is eligible for all programs? Think through your reasoning out loud. Then answer with True or False.

\subsection{``Predict Benefits Reasoning'' for CoT Prompt}

Eligibility: \{eligibility\_requirements\}. 

Predict the programs for which any member of the user's household is eligible. Return only a boolean array of length \{num\_programs\}, e.g. \{example\_array\}, where the value at index `i` is true iff the user is eligible for program `i`. Only return the array. Do not return anything else in the response. If a user's eligibility is unclear, make your best guess. Think through your reasoning out loud.

\subsection{``Predict Benefits Constrained'' for CoT Prompt}

Reasoning: \{reasoning\}. 

Using the reasoning above, predict the programs for which any member of the user's household is eligible. Output a boolean array of length \{num\_programs\}, e.g. \{example\_array\}, where the value at index `i` is true iff the user is eligible for program `i`. If a user's eligibility is unclear, make your best guess.

\subsection{``Predict Clarifying Questions'' for ReAct Chain-of-Thought Prompt}

Eligibility: \{eligibility\_requirements\}. 

Ask a clarifying question that will help you determine if any member of the user's household is eligible for benefits as efficiently as possible. Only ask about one fact at a time. Think through your reasoning out loud, then state your question after a colon, e.g. Question: What is the user's age?

\subsection{``Generate Checker'' Prompt}

\{attempt\_no\}

Eligibility Requirements:  
\{eligibility\_requirement\}

Write a python function called \texttt{check\_eligibility} that takes a dictionary \texttt{hh} containing relevant information and determines user eligibility. \texttt{hh} is a special dictionary connected to a language model that is conversing with the user. Any time it does not contain a key, it will determine that information from the user. As a result here are some requirements for interacting with \texttt{hh}:

\begin{itemize}
    \item DO NOT use \texttt{dict.get()} anywhere in the code. Key errors will be handled elsewhere.
    \item Do not use default values.
    \item Do not use any f-strings, curly brackets, or dynamically generated strings in your keys.
    \item Use only literal strings in keys.
    \item Do not use try-except blocks.
    \item If you need to access data for individuals (rather than the household as a whole) you can use integer indexing. \texttt{hh[0]} is the head of the household. 
\end{itemize}

\texttt{check\_eligibility} returns a bool. All keys and values of \texttt{hh} are strings. If you write helper functions, keep them inside the \texttt{check\_eligibility} function. Make your code as detailed as possible capturing every edge case. Remember that the household may have no relevant members, so be sure to ask about the composition of the household. For example, for childcare programs, check that the household has at least one child. After each new lookup in \texttt{hh}, write a comment suggesting a question to ask. 

The following is a set of preexisting keys and values in the \texttt{hh} dictionary; take care not to duplicate them.

\{preexisting\_keys\}

Avoid using \texttt{int()} and use \texttt{float()} instead. Do not provide anything besides code in your response. Do not use \texttt{input} for user input.

\subsection{``Get Type'' Prompt}

Context:  
\{eligibility\_requirements\}

Code:  
\{code\}

Target key:  
\{key\}

Question: Given the code and context above, what do you expect \{key\} to be an integer, a float, or one choice from a set of strings? Return ONLY int, float, or choice.

\subsection{``Get Values'' Prompt}

Context:  
\{eligibility\_requirements\}

Code:  
\{code\}

Target key:  
\{key\}

Question: Given the code and context above, what are the possible values of \{key\}? Return ONLY the list of possible values in a list of strings. For example, return \texttt{["a", "b", "c"]}.

\subsection{``Extract Values from Answer'' Prompt}

Context:  
\{eligibility\_requirements\}

Line:  
\verb|```{line}```|

We need to extract the value of \{key\} from the following dialog:

Question: \{cq\}  
Answer: \{answer\}

What should we set as the value of \{key\}? Return ONLY the value.

\subsection{``Key Error'' Prompt}
\label{appendix:key-error}

Context:  
\{eligibility\_requirements\}

Line:  
\verb|```{line}```|

We need to determine what value of \{key\} should be stored in the \texttt{hh} dictionary. Ask a question to the user that would get this value. For example, for \texttt{age\_i}, ask ``What is the age of person i?''. Return ONLY the question.

\begin{table*}[!htp]\centering
\scriptsize
\begin{tabular}{lrrrrrrrrr}\toprule
\textbf{Benefits Program} & \textbf{Positive Count} & \textbf{Negative Count} & \textbf{Percentage True (\%)} \\\midrule
FamilyHomelessnessAndEvictionPreventionSupplement & 4 & 3 & 57.14 \\
WorkforceoneCareerCenters & 1 & 1 & 50.00 \\
SilverCorps & 1 & 1 & 50.00 \\
AdultProtectiveServices & 1 & 2 & 33.33 \\
DisabilityRentIncreaseExemption & 8 & 7 & 53.33 \\
ChildTaxCredit & 1 & 4 & 20.00 \\
SeniorCitizenHomeownersExemption & 1 & 9 & 10.00 \\
InfantToddlerPrograms & 3 & 6 & 33.33 \\
LearnEarn & 7 & 1 & 87.50 \\
DisabledHomeownersExemption & 1 & 6 & 14.29 \\
PreKForAll & 1 & 1 & 50.00 \\
JobsPlus & 1 & 1 & 50.00 \\
HeadStart & 6 & 2 & 75.00 \\
KindergartenAndElementarySchool & 1 & 1 & 50.00 \\
CoolingAssistanceBenefit & 4 & 5 & 44.44 \\
HomeEnergyAssistanceProgram & 4 & 3 & 57.14 \\
VeteransAffairsSupportedHousing & 1 & 1 & 50.00 \\
NYCFreeTaxPrep & 2 & 1 & 66.67 \\
FamilyPlanningBenefitProgram & 1 & 6 & 14.29 \\
ChildrenAndYouthWithSpecialHealthCareNeeds & 3 & 2 & 60.00 \\
EnhancedSchoolTaxReliefProgram & 0 & 2 & 0.00 \\
SummerMeals & 1 & 1 & 50.00 \\
TrainEarn & 6 & 3 & 66.67 \\
NYCFinancialEmpowermentCenters & 1 & 1 & 50.00 \\
NYCHAPublicHousing & 3 & 5 & 37.50 \\
ChildAndDependentCareTaxCredit & 4 & 4 & 50.00 \\
ChildCareVouchers & 8 & 2 & 80.00 \\
HIVAIDSServicesAdministration & 1 & 1 & 50.00 \\
BigAppleConnect & 1 & 1 & 50.00 \\
OfficeOfChildSupportServices & 1 & 3 & 25.00 \\
BeaconPrograms & 1 & 1 & 50.00 \\
SafeAndSickLeave & 1 & 5 & 16.67 \\
NYSUnemploymentInsurance & 1 & 1 & 50.00 \\
FamilyTypeHomesForAdults & 1 & 6 & 14.29 \\
EarnedIncomeTaxCredit & 1 & 6 & 14.29 \\
Homebase & 1 & 1 & 50.00 \\
HomeFirstDownPaymentAssistance & 2 & 1 & 66.67 \\
HighSchool & 1 & 1 & 50.00 \\
SeniorCitizenRentIncreaseExemption & 1 & 1 & 50.00 \\
AccessARideParatransitService & 2 & 1 & 66.67 \\
TextTwoWork & 4 & 1 & 80.00 \\
TheEarlyInterventionProgram & 1 & 1 & 50.00 \\
EarlyHeadStart & 2 & 4 & 33.33 \\
Lifeline & 6 & 1 & 85.71 \\
IDNYC & 1 & 1 & 50.00 \\
NYSPaidFamilyLeave & 2 & 1 & 66.67 \\
COVIDnineteenFuneralAssistance & 1 & 1 & 50.00 \\
SchoolAgeAndEarlyChildhoodFamilyAndCommunityEngagementFACECenters & 1 & 1 & 50.00 \\
FairFaresNYC & 1 & 2 & 33.33 \\
NYCYouthHealth & 1 & 1 & 50.00 \\
NewbornHomeVisitingProgram & 4 & 2 & 66.67 \\
AcceleratedStudyInAssociatePrograms & 1 & 1 & 50.00 \\
STEMMattersNYC & 1 & 1 & 50.00 \\
CommoditySupplementalFoodProgram & 1 & 2 & 33.33 \\
CareerAndTechnicalEducation & 1 & 1 & 50.00 \\
NYCHAResidentEconomicEmpowermentAndSustainability & 1 & 1 & 50.00 \\
OutpatientTreatmentServices & 1 & 1 & 50.00 \\
CUNYFatherhoodAcademy & 1 & 3 & 25.00 \\
SummerYouthEmploymentProgram & 1 & 1 & 50.00 \\
ThreeK & 1 & 1 & 50.00 \\
MedicaidForPregnantWomen & 1 & 2 & 33.33 \\
ActionNYC & 1 & 1 & 50.00 \\
FamilyResourceCenters & 2 & 1 & 66.67 \\
NYCCare & 1 & 1 & 50.00 \\
PrimaryAndPreventiveHealthCare & 1 & 1 & 50.00 \\
NYCTenantResourcePortal & 1 & 1 & 50.00 \\
OlderAdultEmploymentProgram & 1 & 1 & 50.00 \\
NYCLaddersForLeaders & 1 & 1 & 50.00 \\
CornerstonePrograms & 1 & 1 & 50.00 \\
ComprehensiveAfterSchoolSystemOfNYC & 1 & 1 & 50.00 \\
WeSpeakNYC & 1 & 1 & 50.00 \\
NYCMitchellLama & 1 & 0 & 100.00 \\
CUNYStart & 1 & 1 & 50.00 \\
NYCNurseFamilyPartnership & 1 & 1 & 50.00 \\
MiddleSchool & 1 & 1 & 50.00 \\
AdvanceEarn & 1 & 1 & 50.00 \\
SectionEightHousingChoiceVoucherProgram & 1 & 0 & 100.00 \\
NYCYouthLeadershipCouncils & 1 & 1 & 50.00 \\
ChildHealthPlusAndChildrensMedicaid & 1 & 2 & 33.33 \\
VeteransPropertyTaxExemption & 1 & 1 & 50.00 \\
FamilyAssessmentProgram & 1 & 1 & 50.00 \\
BasicSchoolTaxReliefProgram & 1 & 0 & 100.00 \\
\hline
\textbf{Total} & \textbf{146} & \textbf{159} & \textbf{47.88} \\
\hline
\end{tabular}
\caption{Benefits Program-wise Positive/Negative Counts and Percentages for the Diversity Dataset}
\end{table*}

\begin{table*}[!htp]\centering
\scriptsize
\begin{tabular}{lrrrrrrrrr}\toprule
\textbf{Benefits Program} & \textbf{Positive Count} & \textbf{Negative Count} & \textbf{Percentage True (\%)} \\\midrule
AdultProtectiveServices & 0 & 3 & 0.00 \\
HomeEnergyAssistanceProgram & 0 & 3 & 0.00 \\
MiddleSchool & 0 & 3 & 0.00 \\
NYCFreeTaxPrep & 0 & 3 & 0.00 \\
NYSPaidFamilyLeave & 1 & 2 & 33.33 \\
NYSUnemploymentInsurance & 0 & 3 & 0.00 \\
WorkforceoneCareerCenters & 1 & 2 & 33.33 \\
FamilyTypeHomesForAdults & 2 & 1 & 66.67 \\
HeadStart & 0 & 3 & 0.00 \\
NYCFinancialEmpowermentCenters & 3 & 0 & 100.00 \\
TextTwoWork & 3 & 0 & 100.00 \\
ThreeK & 0 & 3 & 0.00 \\
CornerstonePrograms & 0 & 3 & 0.00 \\
JobsPlus & 0 & 3 & 0.00 \\
CommoditySupplementalFoodProgram & 0 & 3 & 0.00 \\
NYCCare & 0 & 3 & 0.00 \\
SilverCorps & 0 & 3 & 0.00 \\
SummerMeals & 1 & 2 & 33.33 \\
PrimaryAndPreventiveHealthCare & 1 & 2 & 33.33 \\
IDNYC & 3 & 0 & 100.00 \\
NYCYouthLeadershipCouncils & 0 & 3 & 0.00 \\
Homebase & 1 & 2 & 33.33 \\
NYCMitchellLama & 2 & 1 & 66.67 \\
NYCNurseFamilyPartnership & 0 & 3 & 0.00 \\
AdvanceEarn & 0 & 3 & 0.00 \\
BeaconPrograms & 1 & 2 & 33.33 \\
ChildHealthPlusAndChildrensMedicaid & 0 & 3 & 0.00 \\
BasicSchoolTaxReliefProgram & 0 & 3 & 0.00 \\
TheEarlyInterventionProgram & 0 & 3 & 0.00 \\
AccessARideParatransitService & 3 & 0 & 100.00 \\
ChildAndDependentCareTaxCredit & 0 & 3 & 0.00 \\
FamilyResourceCenters & 0 & 3 & 0.00 \\
InfantToddlerPrograms & 0 & 3 & 0.00 \\
NYCTenantResourcePortal & 3 & 0 & 100.00 \\
NYCYouthHealth & 1 & 2 & 33.33 \\
DisabledHomeownersExemption & 0 & 3 & 0.00 \\
OutpatientTreatmentServices & 0 & 3 & 0.00 \\
STEMMattersNYC & 0 & 3 & 0.00 \\
SeniorCitizenHomeownersExemption & 0 & 3 & 0.00 \\
CareerAndTechnicalEducation & 0 & 3 & 0.00 \\
NewbornHomeVisitingProgram & 0 & 3 & 0.00 \\
SchoolAgeAndEarlyChildhoodFamilyAndCommunityEngagementFACECenters & 0 & 3 & 0.00 \\
BigAppleConnect & 0 & 3 & 0.00 \\
CUNYFatherhoodAcademy & 0 & 3 & 0.00 \\
HomeFirstDownPaymentAssistance & 0 & 3 & 0.00 \\
DisabilityRentIncreaseExemption & 0 & 3 & 0.00 \\
KindergartenAndElementarySchool & 0 & 3 & 0.00 \\
EarnedIncomeTaxCredit & 1 & 2 & 33.33 \\
HIVAIDSServicesAdministration & 0 & 3 & 0.00 \\
OlderAdultEmploymentProgram & 0 & 3 & 0.00 \\
FamilyHomelessnessAndEvictionPreventionSupplement & 0 & 3 & 0.00 \\
ChildCareVouchers & 1 & 2 & 33.33 \\
ComprehensiveAfterSchoolSystemOfNYC & 1 & 2 & 33.33 \\
COVIDnineteenFuneralAssistance & 0 & 3 & 0.00 \\
TrainEarn & 0 & 3 & 0.00 \\
LearnEarn & 0 & 3 & 0.00 \\
SectionEightHousingChoiceVoucherProgram & 0 & 3 & 0.00 \\
CoolingAssistanceBenefit & 0 & 3 & 0.00 \\
MedicaidForPregnantWomen & 0 & 3 & 0.00 \\
SummerYouthEmploymentProgram & 1 & 2 & 33.33 \\
FairFaresNYC & 0 & 3 & 0.00 \\
PreKForAll & 0 & 3 & 0.00 \\
ChildrenAndYouthWithSpecialHealthCareNeeds & 1 & 2 & 33.33 \\
CUNYStart & 2 & 1 & 66.67 \\
NYCLaddersForLeaders & 0 & 3 & 0.00 \\
FamilyAssessmentProgram & 1 & 2 & 33.33 \\
FamilyPlanningBenefitProgram & 0 & 3 & 0.00 \\
NYCHAPublicHousing & 0 & 3 & 0.00 \\
SafeAndSickLeave & 2 & 1 & 66.67 \\
WeSpeakNYC & 1 & 2 & 33.33 \\
VeteransAffairsSupportedHousing & 0 & 3 & 0.00 \\
NYCHAResidentEconomicEmpowermentAndSustainability & 0 & 3 & 0.00 \\
SeniorCitizenRentIncreaseExemption & 0 & 3 & 0.00 \\
AcceleratedStudyInAssociatePrograms & 0 & 3 & 0.00 \\
EnhancedSchoolTaxReliefProgram & 0 & 3 & 0.00 \\
EarlyHeadStart & 0 & 3 & 0.00 \\
ActionNYC & 0 & 3 & 0.00 \\
Lifeline & 1 & 2 & 33.33 \\
VeteransPropertyTaxExemption & 0 & 3 & 0.00 \\
HighSchool & 0 & 3 & 0.00 \\
OfficeOfChildSupportServices & 0 & 3 & 0.00 \\
ChildTaxCredit & 0 & 3 & 0.00 \\
\hline
\textbf{Total} & \textbf{38} & \textbf{208} & \textbf{15.42} \\
\hline
\end{tabular}
\caption{Benefits Program-wise Positive/Negative Counts and Percentages for the Representative Dataset}
\end{table*}

\newpage

\section{Instructions to User Survey Participants}

Please select five programs in which you are interested. The chatbot will help you determine your eligibility for them. Please answer the questions with respect to your own household. You may end the study at any time. Your responses will not be stored. Only the final model prediction and your feedback will be used.

Participants were students and a New York City university recruited in person and compensated up to \$5 USD.

\section{Constrained Generation Implementation Details}

For constrained generation, when available (e.g., for OpenAI models) we use tool use JSON formatting to ensure constraints are satisfied.
Where the constrains cannot be clearly specified in the API, we check whether the output conforms to a validation function (e.g., with regular expressions) and regenerate it if it does not.
For hugging face models, we use the Outlines library \texttt{choices} and \texttt{regex} functions in most cases \cite{willard2023efficient}.
For ReAct-style prompting, we first instruct the model to think through their reasoning. Then, in a second prompt, we prompt the model to output its answer in the specified form based on the thoughts.

\end{document}